%% file: aaai25.tex
\documentclass[letterpaper]{article} 
\usepackage{aaai25}  
\usepackage{times}  
\usepackage{helvet}  
\usepackage{courier}  
\usepackage[hyphens]{url}  
\usepackage{graphicx} 
\usepackage{natbib}  
\usepackage{caption} 
\usepackage{algorithm}
\usepackage{algorithmic}

\usepackage{amsmath,amsfonts}
\usepackage{array}
\usepackage{subfig}
\usepackage{textcomp}
\usepackage{url}
\usepackage{verbatim}
\usepackage{graphicx}
\usepackage{cite}
\usepackage{xcolor}
\usepackage{soul}
\usepackage{amssymb}
\usepackage{enumitem}

\usepackage{arydshln}
\usepackage{fancyhdr}
\usepackage{varwidth}
\usepackage{arydshln}
\usepackage{multirow}
\usepackage{mathtools}
\usepackage{algorithmic}
\usepackage{algorithm}

\usepackage{datetime2}

\usepackage{newfloat}
\usepackage{listings}

\usepackage{kotex}
\usepackage{soul}
\usepackage{pifont}
\newcommand{\cmark}{\ding{51}}%
\newcommand{\xmark}{\ding{55}}%

\newcommand{\sys}{\textsf{AT-SNN}}
\newcommand{\syss}{\textsf{AT-SNN$^\textsf{e}$}}

\definecolor{caribbeangreen}{rgb}{0.0, 0.6, 0.6}
\newcommand{\kdh}[1]{{\color{orange}{#1}}}

\newcommand{\remove}[1]{}

\newcommand{\ym}[1]{{\color{green}{#1}}}

\DeclareCaptionStyle{ruled}{labelfont=normalfont,labelsep=colon,strut=off} 
\floatstyle{ruled}
\newfloat{listing}{tb}{lst}{}
\floatname{listing}{Listing}
\nocopyright
\title{\sys{}: Adaptive Tokens for Vision Transformer on Spiking Neural Network}
\author {
    Donghwa Kang\textsuperscript{\rm 1},
    Youngmoon Lee\textsuperscript{\rm 4},
    Eun-Kyu Lee\textsuperscript{\rm 5},
    Brent Byunghoon Kang\textsuperscript{\rm 1},
    Jinkyu Lee\textsuperscript{\rm 3},
    Hyeongboo Baek\textsuperscript{\rm 2}
}
\affiliations {
    \textsuperscript{\rm 1}KAIST  
    \textsuperscript{\rm 2}University of Seoul  
    \textsuperscript{\rm 3}Sungkyunkwan University  
    \textsuperscript{\rm 4}Hanyang University  
    \textsuperscript{\rm 5}Incheon National University  
}

\usepackage{bibentry}

\begin{document}

\maketitle

\input{00abstract}

%
\input{01introduction}

\input{02related_works}

\input{04_method}

\input{05evaluation}

\input{06conclusion}

\appendix

\bigskip

\bibliography{aaai25}

\end{document}

%% file: 00abstract.tex
\begin{abstract}
In the training and inference of spiking neural networks (SNNs), direct training and lightweight computation methods have been orthogonally developed, aimed at reducing power consumption.
However, only a limited number of approaches have applied these two mechanisms simultaneously and failed to fully leverage the advantages of SNN-based vision transformers (ViTs) since they were originally designed for convolutional neural networks (CNNs).
In this paper, we propose \sys{} designed to dynamically adjust the number of tokens processed during inference in SNN-based ViTs with direct training by considering power consumption proportional to the number of tokens.
We first demonstrate the applicability of adaptive computation time (ACT), previously limited to RNNs and ViTs, to SNN-based ViTs, enhancing it to discard less informative spatial tokens selectively.
Also, we propose a new token-merge mechanism that relies on the similarity of tokens, which further reduces the number of tokens while enhancing accuracy.  
We implement \sys{} to Spikformer and show the effectiveness of \sys{} in achieving high energy efficiency and accuracy compared to state-of-the-art approaches on the image classification tasks, CIFAR-10, CIFAR-100, and TinyImageNet. 
Notably, our approach uses up to 42.4\% fewer tokens than the existing best-performing method on CIFAR-100, while conserving higher accuracy.

\end{abstract}

%% file: 01introduction.tex
\section{Introduction}
\label{sec:intro}


As an alternative to the high energy consumption of artificial neural networks (ANNs), spiking neural networks (SNNs) have recently received considerable attention~\cite{STO18,NCC22,LDB20}. 
SNNs, regarded as the next generation of neural networks, emulate the synaptic functioning of the human brain. 
While ANNs perform a floating point operation on a single timestep, SNNs operate by repeatedly computing binary-formatted spikes over multiple timesteps, allowing them to achieve accuracy comparable to ANNs with lower power consumption. 

\begin{figure}[t!] 
    \centering
    \includegraphics[width=1\linewidth]{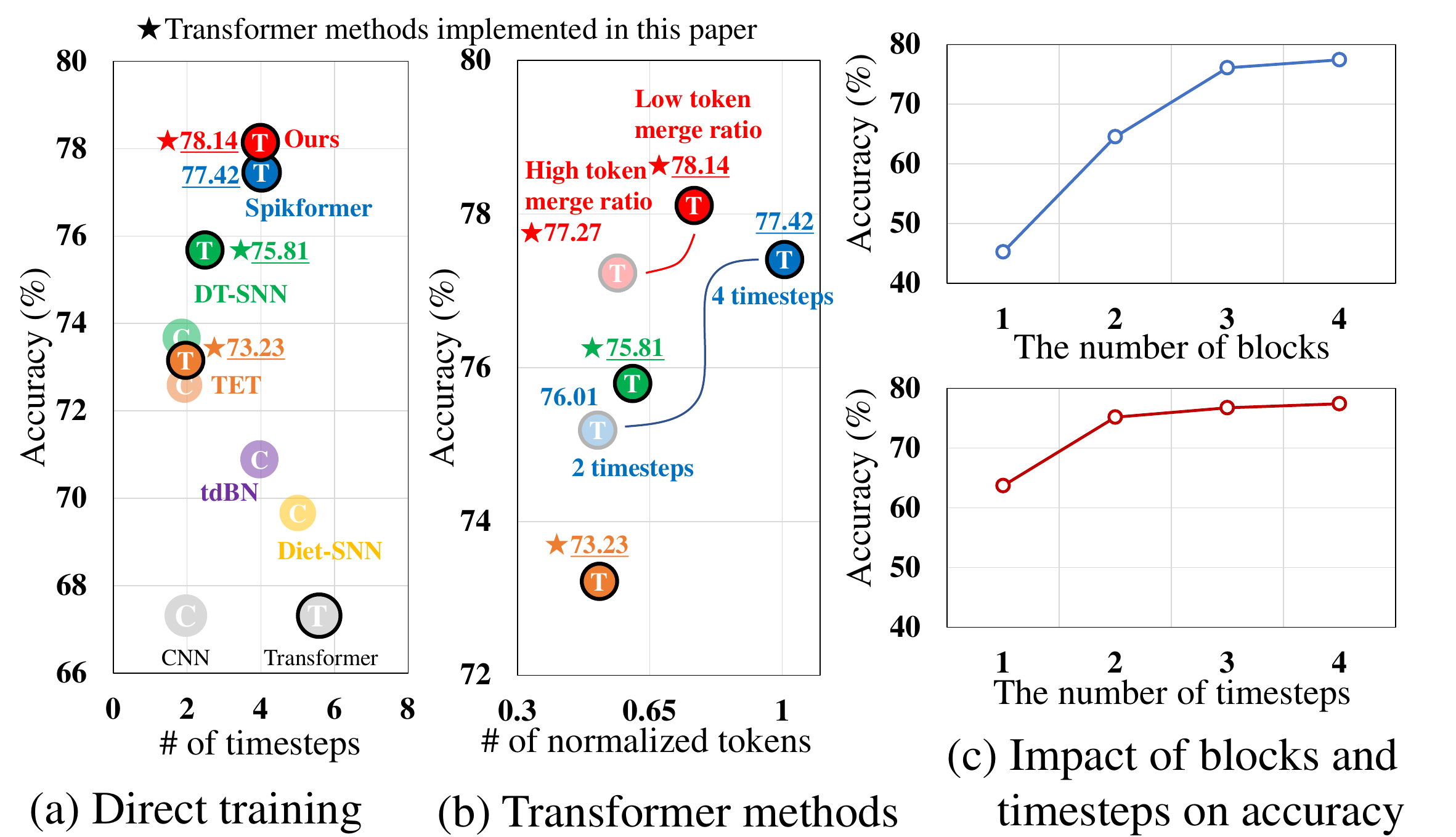}
    \caption{Accuracy comparison of lightweight computation methods in direct training on CIFAR-100.}
    \label{fig:intro}
    \vspace{-0.2cm}
\end{figure}

State-of-the-art research aimed at reducing power consumption in SNNs has led to significant advancements in both training and inference.
Initially, training methods have progressed from the ANN-to-SNN~\cite{SYW19}, which necessitates hundreds of timesteps, to direct training~\cite{WDL18} that requires substantially fewer timesteps with a marginal accuracy drop~\cite{TLX23,WKZ22}. 
Regarding inference, lightweight computation methods have been proposed to provide the trade-off between timesteps and accuracy. 
However, only a few approaches apply direct training and lightweight computation methods simultaneously, and these were initially designed for convolutional neural network (CNN) models, e.g., the circled `C' of Diet-SNN~\cite{RAR20}, tdBN~\cite{ZWD21}, TET~\cite{DLZ22}, and DT-SNN~\cite{LMG23} in Fig.~\ref{fig:intro}(a).
Consequently, they fail to fully harness the potential of the vision transformer (ViT) models, which typically provide higher accuracy than CNNs, e.g., the circled `T' of Spikformer~\cite{ZZH22} with 77.42\% compared to the circled `C' of other methods in Fig.~\ref{fig:intro}(a).

Unlike CNNs, ViTs segment the input image into small patches, treating each as an independent token and executing spiking self-attention (SSA) to assess token relationships. 
This operation requires substantial computational resources and, consequently, accounts for the majority of energy consumption in ViT~\cite{ZZH22}.
Our strategy is to employ adaptive computation time (ACT)~\cite{GRA16} to improve its implementation in SNN-based ViTs, enabling the selective omission of spatial tokens that are less informative with direct training.
ACT, initially applied to recurrent neural networks (RNNs), to model neural outputs with a halting distribution, converts the discrete halting issue into a continuous optimization problem, thereby minimizing total computation~\cite{GRA16}. 
Subsequently, ACT was applied to ViTs, by variably masking tokens based on their halting probability, focusing on the most informative tokens. 
However, the applicability of ACT to SNN-based ViT has not been identified yet.

In this paper, we propose \sys{}, a method designed to dynamically adjust the number of tokens processed during network inference in SNN-based ViTs with direct training. 
We first demonstrate the applicability of ACT to SNN, previously limited to RNNs and ViTs, to SNN-based ViTs, enhancing it to discard less informative spatial tokens selectively (to be detailed in~\ref{subsec:motivation}). 
Existing lightweight computation methods that do not consider SNN primarily perform block-level halting (e.g., A-ViT~\cite{YVA22}), based on the observation that accuracy converges as the number of blocks involved in inference increases (as shown in the upper subfigure of Fig.~1(c)). 
Conversely, techniques that take into account the characteristics of SNN perform timestep-level halting (e.g., DT-SNN~\cite{LMG23}), based on the observation that accuracy converges as the number of timesteps increases (as shown in the lower subfigure of Fig.~1(c)). 
\sys{} leverages both of these approaches to perform two-dimensional halting (to be detailed in~\ref{subsec:AT-SNN}), which is the first attempt in SNN, at least for direct training.
Additionally, we propose a new token-merge mechanism incorporated into \sys{} that relies on token similarity, further reducing the number of tokens while enhancing accuracy.  

We implement \sys{} in Spikformer and demonstrate its effectiveness in achieving high energy efficiency and accuracy compared to state-of-the-art approaches on the image classification tasks, CIFAR-10, CIFAR-100~\cite{KHO09}, and TinyImageNet~\cite{LYY15}.\footnote{For a fair comparison, we implemented TET and DT-SNN (initially designed for CNNs) into Spikformer.} 
For example, as shown in Fig.~\ref{fig:intro}(a), \sys{} significantly outperforms Spikformer-based TET and DT-SNN in terms of accuracy (e.g., 73.23\% and 75.81\%, respectively). 
The proposed token-merge mechanism provides a trade-off between the number of tokens involved in inference and accuracy. 
As shown in Fig.~\ref{fig:intro}(b), \sys{} achieves higher accuracy with fewer tokens compared to Spikformer executing four timesteps (and two timesteps) at a low (and high) token merge ratio.


\begin{figure}[t!]
    \centering
    \includegraphics[width=1\linewidth]{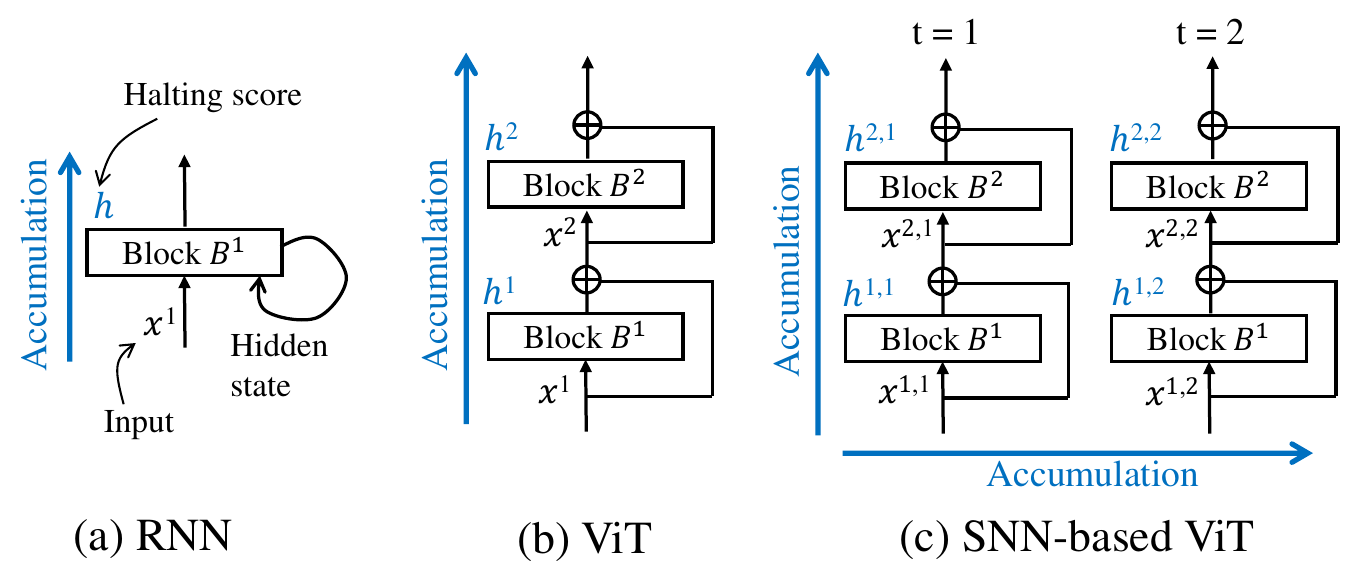}
    \caption{Comparison of model architecture and halting-score accumulation paths 
    among RNN, ViT, and SNN-based ViT when ACT is applied.}
    \label{fig:ACT_multidimension}
     \vspace{-0.2cm}
\end{figure}

Our contribution can be summarized as follows:
\begin{itemize}
    \item We first demonstrate the applicability of ACT to SNN with direct training.
    \item We propose \sys{}, a two-dimensional halting method incorporating ACT into SNN-based ViT.
    \item We propose a token-merge mechanism applied to \sys{} further reducing the number of tokens while enhancing accuracy. 
    \item We conducted experiments on CIFAR-10, CIFAR-100, and TinyImageNet, demonstrating efficiency in accuracy and power consumption compared to state-of-the-art methods.
\end{itemize}

%% file: 02related_works.tex
\section{Related Work}
\label{sec:related_work}

\begin{figure}[t!]
    \centering
    \includegraphics[width=1\linewidth]{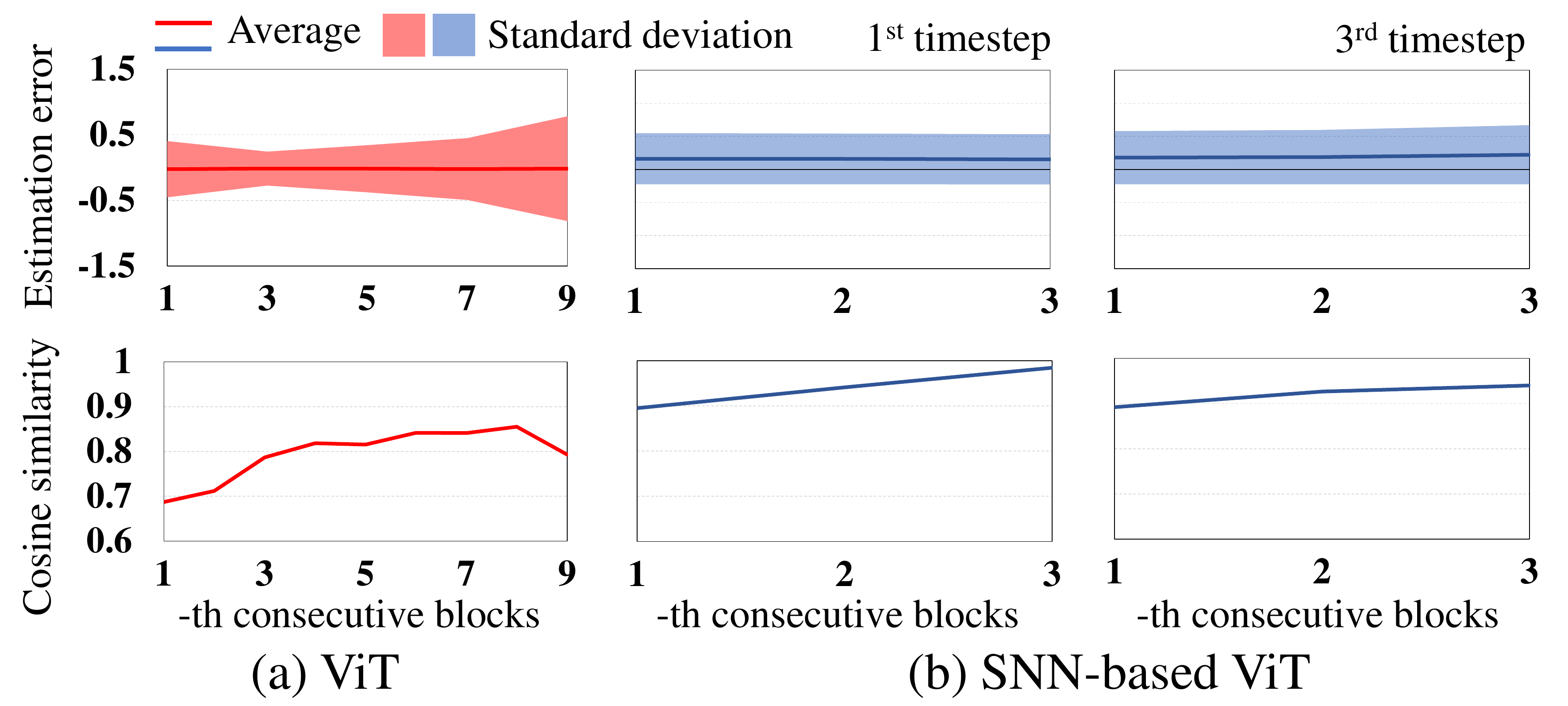}
    \caption{Estimation error and cosine similarity of tokens between consecutive blocks for (a) 12-layer ViT and (b) 4-layer SNN-based ViT (Spikformer) on CIFAR-100.}
    \label{fig:similarity}
    \vspace{-0.2cm}
\end{figure}

\begin{figure*}[t]
    \centering
    \includegraphics[width=1\linewidth]{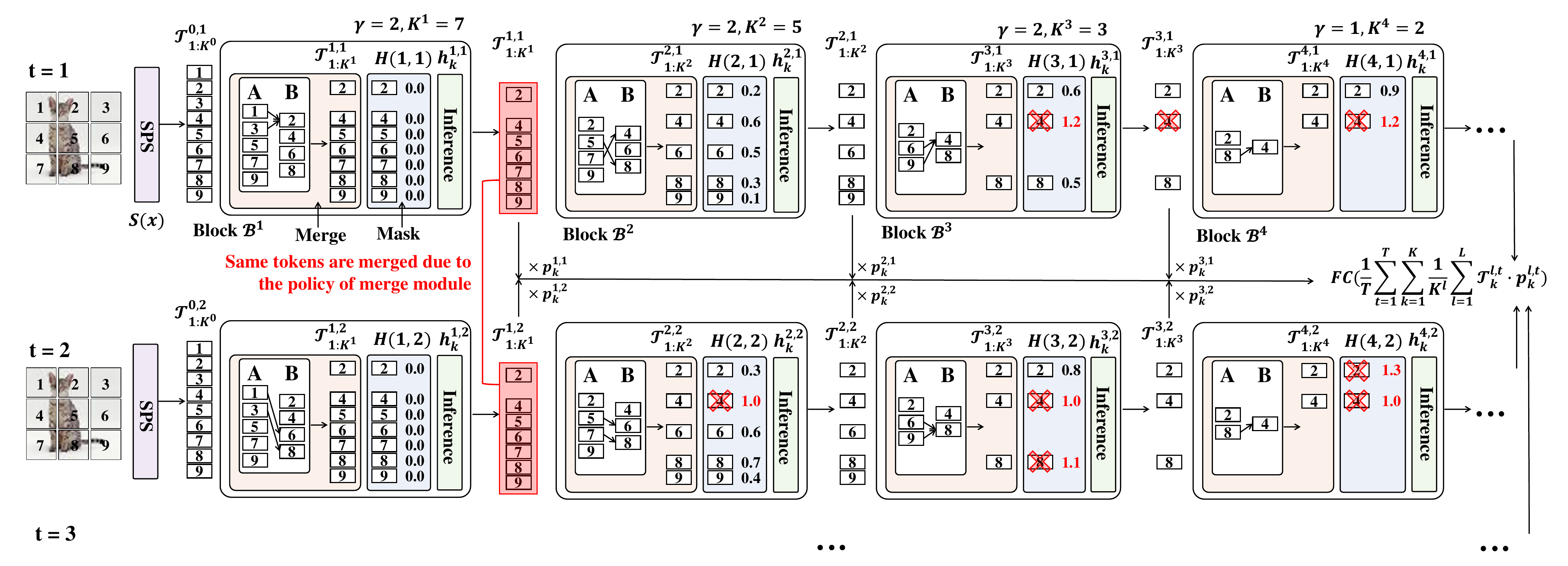}
    \caption{Token-level merging and masking example of \sys{}: At the first timestep $t=1$, the input $x$ passes through the SPS, generating a token set $\mathcal{T}^{l, t}_{1:K^{0}}$. In the first block $\mathcal{B}^1$, with $\gamma = 2$ for nine tokens, the first and third tokens are merged into the second token by the merge module, and halting scores $h_k^{1,1}$ are added through inference. In subsequent blocks, tokens are merged based on their respective $\gamma$ values, and tokens with accumulated halting scores $H(l,t)$ of one or greater are masked.
    From the second timestep onwards, the same operations are repeated on the same input $x$. The halting score accumulation follows Eq.~\eqref{eq:M_score}, and the merged tokens within the same block across timesteps remain consistent due to the merge module's policy (detailed in Algo.~\ref{algo:merge} and Sec.~\ref{subsec:eval_ablation}). The vector values of merged or masked tokens are set to zero, and no further halting score is accumulated for the tokens.
    For ease of implementation, a masked token can also be considered a candidate for merging. 
    For simplicity, this example does not include timestep-level halting score accumulation.}
    \label{fig:main_example}
    \vspace{-0.2cm}
\end{figure*}

Methods like DT-SNN and SEENN~\cite{LMG23, LGK24} dynamically adjust timesteps of SNN during inference based on accuracy needs, using entropy and confidence metrics. 
SEENN employs reinforcement learning to optimize timesteps for each image, while TET~\cite{DLZ22} introduces a loss function to address gradient loss in spiking neurons, achieving higher accuracy with fewer timesteps. 
However, these methods are less suitable for deeper models requiring fewer timesteps, where efficiency gains are limited. 
MST~\cite{WFC23} proposes an ANN-to-SNN conversion method for SNN-based ViTs, using token masking within model blocks to reduce energy consumption by decreasing spiking tokens during inference. 
Despite its effectiveness, MST still requires hundreds of timesteps by relying on ANN-to-SNN.

ACT~\cite{GRA16} dynamically allocates inference time for RNN models based on the difficulty of the input, enhancing accuracy in natural language processing (NLP) tasks. 
SACT~\cite{FCZ17} adapts ACT for ResNet architectures, allowing the model to halt inference early depending on the input data, thus maintaining classification accuracy while reducing FLOPs. 
Similarly, A-ViT~\cite{YVA22} dynamically adjusts token computation within transformers to optimize efficiency. 
However, these studies, which are based on ANN, do not account for the timestep characteristics of SNN and typically perform a single inference per input. 
LFACT~\cite{ZEK21} expands ACT to enable repeated inferences across input sequences, though it remains limited to RNNs. 
In contrast, \sys{} must consider multiple timesteps and blocks, making it uniquely suited for SNN-based vision transformers.

%% file: 04_method.tex
\section{Method}
\label{sec:method}

\remove{
\begin{figure}[t]
\begin{minipage}[t]{0.34\columnwidth}
        \includegraphics[width=\linewidth, height=1.1cm]{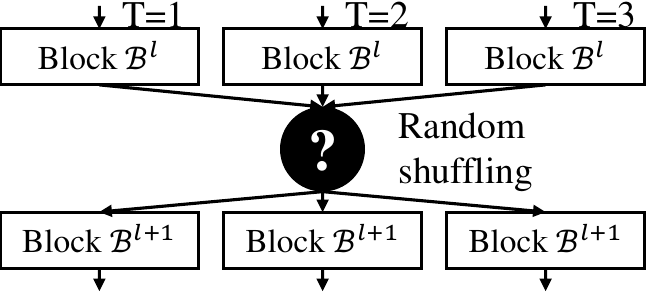}
        \captionof{figure}{Suffling}
        \label{fig:shuffle_and_mask}
\end{minipage}
\hfill
\begin{minipage}[t]{0.65\columnwidth}
\vspace{-1cm} 
    \resizebox{\columnwidth}{!}{%
    \begin{tabular}{c|c|c|c}
    \hline
    Dataset       & Model            & Clean & Shuffled \\ \hline
    Cifar-10      & Spikformer-4-384 & 94.88 & 94.38    \\
    Cifar-100     & Spikformer-4-384 & 77.42 & 76.88    \\
    Tiny-ImageNet & Spikformer-8-384 & 65.23 & 64.33    \\ \hline
    \end{tabular}%
    }
    \captionof{table}{Accuracy }
    \label{tab:shuffle_spike}
\end{minipage}
\end{figure}
}

\subsection{Can ACT be applied to SNN-based ViT?}
\label{subsec:motivation}

ACT was initially proposed for RNNs to provide a condition for accuracy-effective halting during inference. 
As illustrated in Fig.~\ref{fig:ACT_multidimension}(a), an encoder block $\mathcal{B}^1$ of an RNN repeatedly processes the same input $x^1$ while the hidden state evolves.
During training, ACT employs a unique loss function to learn the impact of execution halting for the combination of $\mathcal{B}^1$ and $x^1$ on the accuracy, thereby determining the halting probability, referred to as the halting score $h$. 
During inference, $h$ is repeatedly added, and the halting condition is met when the accumulated halting scores reach one. 



On the other hand, ViT has two distinct properties: (i) ViT has multiple identical encoder blocks as shown in Fig.~\ref{fig:ACT_multidimension}(b);
and (ii) the input vectors between consecutive blocks have low estimation errors and high cosine similarity 
as shown in Fig.~\ref{fig:similarity}(a),  mainly because the input $x^2$ of $\mathcal{B}^2$ is the sum of the output of $\mathcal{B}^1$ and $x^1$ in Fig.~\ref{fig:ACT_multidimension}(a). 
Due to (i) and (ii), the inputs of each block, similar to the initial input $x^1$, accumulate halting scores (e.g., $h^1 + h^2 + \cdots$) as they pass through the identical encoder blocks. 
Similar to ACT in RNNs, the process halts when the accumulated halting scores reach one.


As illustrated in Fig.~\ref{fig:ACT_multidimension}(c), the SNN-based ViT possesses an architectural structure that is fundamentally analogous to that of the standard ViT, yet it performs iterative computations across multiple timesteps. 
The primary distinction lies in the storage of information as membrane potential within each block, which, upon surpassing a specific threshold, triggers a binary activation function that outputs a value of one. 
Consequently, property (i) is satisfied, but it is necessary to substantiate the validity of property (ii).
As depicted in Fig.~\ref{fig:similarity}(b), the inputs between consecutive blocks at a timestep exhibit low estimation errors and high (even higher than the ViT case in Fig.~\ref{fig:similarity}(a)) cosine similarity.

In light of the considerations outlined in points (i) and (ii) of the SNN-based ViT, it is feasible to apply ACT independently for each timestep. 
For instance, at a given timestep $t$, the sum $h^{1,t} + h^{2,t} + \cdots$ is computed solely for that specific timestep as a halting condition. 
However, drawing inspiration from the motivation presented in Fig.~\ref{fig:intro}(c), we propose that accumulating greater halting probabilities (as an execution penalty) to timesteps with higher timestep-index can lead to more accuracy-effective computation. 
For example, during the inference for $\mathcal{B}^{2}$ at the second timestep, \sys{} additionally incorporates the halting score from $\mathcal{B}^{1}$ at the first timestep (i.e., $h^{1,1} + h^{1,2} + h^{2,2}$ according to Eq.~\eqref{eq:M_score}), which will be elaborated in Sec.~\ref{subsec:AT-SNN}.



\subsection{Adaptive Tokens for SNN-based ViT}
\label{subsec:AT-SNN}

We formulate the SNN-based ViT as follows~\cite{ZYZ23}:

\begin{equation}
    f_{T}(x) = FC(\frac{1}{T} \sum^{T}_{t=1} \mathcal{B}^{L} \circ \mathcal{B}^{L-1} \circ \cdot \cdot \cdot \circ \mathcal{B}^{1} \circ \mathcal{S}(x)),
    \label{eq:gradient}
\end{equation}

\noindent
where $x \in \mathbb{R}^{T \times C \times H \times W}$ is the input of which $T$, $C$, $H$, and $W$ denote the timesteps, channels, height, and width. 
The function $\mathcal{S}(\cdot)$ represents the spike patch splitting (SPS) module, which divides the input image into $K^{0}$ tokens. 
The function $\mathcal{B}(\cdot)$ denotes a single encoder block, consisting of spike self-attention (SSA) and a multi-layer perceptron (MLP), with a total of $L$ blocks in the model.
The function $FC(\cdot)$ represents a fully-connected layer. 
Finally, the tokens passing through all blocks are averaged and input to $FC(\cdot)$.

After passing through $\mathcal{S}(x)$ at a timestep $t$, the input image $x$ is divided into a set of $K^{0}$ tokens denoted by $\mathcal{T}^{0,t}_{1:K^{0}}$. 
Let $\mathcal{T}^{l, t}_{1:K^{l}}$ represent the set of $K^{l}$ remaining (i.e., not merged) tokens after processing merge module in the $l$-th (for $l > 0$) block at the $t$-th timestep, which is expressed as follows:

\begin{equation}
    \mathcal{T}^{l, t}_{1:K^{l}} = \mathcal{B}^{l}(\mathcal{T}^{l-1, t}_{1:K^{l-1}}).
    \label{eq:tokenes}
\end{equation}

\noindent
The halting score $h^{l, t}_{k}$ of the $k$-th token at the $t$-th timestep in the $l$-th block can be defined as follows:

\begin{equation}
    h^{l, t}_{k} = \sigma(\alpha \times \frac{\mathcal{T}^{l, t}_{k, 1}}{\mathcal{NT}^{l,t}_{k}} + \beta),
    \label{eq:halting}
\end{equation}

\noindent
where $\sigma(\cdot)$ denotes the logistic sigmoid function, and $\alpha$ and $\beta$ are scaling factors.
Let $\mathcal{T}^{l,t}_{k}$ represent the embedding vector of the $k$-th token, and $\mathcal{T}^{l,t}_{k, 1}$ denote the first element of this vector.
\sys{} merges tokens within each block by summing their values to maintain discrete spikes, a characteristic of SNNs where all embedding vector elements are positive. Consequently, as a token undergoes more merges, $\mathcal{T}^{l,t}_{k, 1}$ increases.
To ensure fair halting scores for merged tokens, we define $\mathcal{NT}^{l,t}_{k}$ as follows:
Initially, $\mathcal{NT}^{l,t}_{k}$ is set to one for all $k$-th tokens. 
When the $i$-th token merges into the $j$-th token, $\mathcal{NT}^{l,t}_{j}$ is updated to $\mathcal{NT}^{l,t}_{j} = \mathcal{NT}^{l,t}_{j} + \mathcal{NT}^{l,t}_{i}$, and $\mathcal{NT}^{l,t}_{i}$ is set to zero. 
Likewise, if the $i$-th token is masked, $\mathcal{NT}^{l,t}_{i}$ is also set to zero.\footnote{Note that a token with $\mathcal{NT}^{l,t}_{i}$ equal to zero no longer accumulates halting scores.}
The sigmoid function ensures that $0 \leq h^{l, t}_{k} \leq 1$. 
\sys{} calculates $h^{l, t}_{k}$ using the first element of the embedding vector of the token 
like A-ViT, and the first node of MLP in each block learns the halting score.

\sys{} accumulates halting scores across blocks within a single timestep and continues to accumulate scores from previous timesteps and blocks over multiple timesteps, as a two-dimensional halting policy.
\sys{} defines the halting module $H_{k}(L', T')$ at the $T'$-th timestep and the $L'$-th block as follows:

\begin{gather}
    H_{k}(L', T') = \sum^{L'-1}_{l=1}\sum^{T'}_{t=1}h^{l, t}_{k}.
    \label{eq:M_score}
\end{gather}

\noindent
\sys{} masks tokens with $H_{k}(L', T') \geq 1 - \epsilon$ in each block.  
If the 
$k$-th token is masked at the $L'$-th block and $T'$-th timestep, it remains zeroed out from the $L'+1$ block onward in the $T'$-th timestep.
Fig.~\ref{fig:main_example} illustrates a token-level merging and masking example of AT-SNN.

Based on the defined halting score, we propose a new loss function that allows \sys{} to determine the required number of tokens according to the input image during training. 
We define $\mathcal{N}^{t}_{k}$ as the index of the block where the $k$-th token halts at the $t$-th timestep, which is obtained by 

\begin{equation}
    \mathcal{N}^{t}_{k} = \underset{l \leq L}{\mathrm{arg\,min}}\ H_{k}(l, t) \geq 1 - \epsilon,
\end{equation}

\begin{algorithm} [t]
\caption{Token merge algorithm for $\mathcal{B}^{l}$ at $t$}
\label{algo:merge}
\small
\textbf{Input:} input token set $\mathcal{T}^{l-1,t}_{1:K^{l-1}}$ and a set of merged tokens $\mathbb{A}^{l}$ in $\mathcal{B}^{l}$ at $t$ (for $t=1$, $\mathbb{A}=\emptyset$) 
\begin{algorithmic}[1]
    \STATE $\gamma \leftarrow \min(\gamma, \text{floor}(K^{l-1} / 2))$
    \IF{t == 1}
        \STATE $A \leftarrow$ even-numbered tokens in $\mathcal{T}^{l-1,t}_{1:K^{l-1}}$
    \ELSE
        \STATE $A \leftarrow \mathbb{A}^{l}$
    \ENDIF
    \STATE $B \leftarrow$ odd-numbered tokens in $\mathcal{T}^{l-1,t}_{1:K^{l-1}}$    
    \FOR{each iteration up to $\gamma$}
        \STATE For $\mathcal{T}_i^{l,t} \in A$ and $\mathcal{T}_j^{l,t} \in B$ with the highest cosine similarity,  $\mathcal{T}_i^{l,t}$ is merged to $\mathcal{T}_j^{l,t}$.
        \STATE $A \leftarrow A \setminus \{\mathcal{T}_i^{l,t}\}$
    \ENDFOR
    \IF {$\mathbb{A}^{l} == \emptyset$}
        \STATE $\mathbb{A}^{l} \leftarrow$ a set of merged tokens in $A$ to $B$.
    \ENDIF
    
    \RETURN $\mathcal{T}^{l,t}_{1:K^{l}}$ and $\mathbb{A}^{l}$ 
\end{algorithmic}
\normalsize
\end{algorithm}

\noindent
where $\epsilon$ is a constant value that determines the threshold for the halting score. 
Additionally, we define an auxiliary variable, remainder, to track the remaining amount of halting score for each token until it halts at each timestep and layer as follows:

\begin{equation}
    r^{l,t}_{k} = 1 - H_{k}(l-1, t).
\end{equation}

\noindent
Then, we define the halting probability of each token at each timestep and block as follows.

\begin{equation}
    p^{l, t}_{k} = \Bigg\{ \begin{array}{l}
        h^{l,t}_{k}\text{ if }t = \{1, ..., T\}\text{ and }l < \mathcal{N}^{t}_{k}\\
        r^{l,t}_{k}\text{ if }t = \{1, ..., T\}\text{ and }l = \mathcal{N}^{t}_{k}\\
        0\text{ otherwise }
    \end{array}
\end{equation}


\noindent
According to the definitions of $h^{l,t}_{k}$ and $r^{l,t}_{k}$, $0 \leq p^{l, t}_{k} \leq 1$ holds. 

Based on the previously defined halting module and probability, we propose the following loss functions for training \sys{}. First, we apply a mean-field formulation (halting-probability weighted average of previous states) to the output at each block and timestep, accumulating the results. Therefore, the classification loss function $\mathcal{L}_{task}$ is defined as follows.

\begin{equation}
    \mathcal{L}_{task} = \mathcal{C}(FC(\frac{1}{T}\sum^{T}_{t=1}\sum^{L}_{l=1}\frac{1}{K^{l}}\sum^{K^{l}}_{k=1}\mathcal{T}^{l, t}_{k} \cdot p^{l, t}_{k})),
    \label{eq:cross_entrophy}
\end{equation}

\noindent
where $\mathcal{C}$ denotes the cross-entropy loss. 
Next, we propose a loss function to encourage each token to halt at earlier timesteps and blocks, using fewer computations. First, we define $\mathcal{NB}^{t}_{k}$, which represents the number of blocks over which the halting score has been accumulated until the token halts at a particular block.

\begin{equation}
    \mathcal{NB}^{t}_{k} = t \times (\mathcal{N}^{t}_{k}-1) + 1
    \label{eq:nb}
\end{equation}

\noindent
Then, we can define the ponder loss $\mathcal{L}_{ponder}$ as follows:

\begin{equation}
    \mathcal{L}_{ponder} = \frac{1}{TK^{L}} \sum_{t=1}^{T}\sum^{K^{L}}_{k=1}(\mathcal{NB}^{t}_{k} + r_{k}^{\mathcal{N}^{t}_{k}, t}).
    \label{eq:ponder_loss}
\end{equation}

\noindent
$\mathcal{L}_{ponder}$ consists of the average number of blocks over which each token accumulates its halting score and the average remainder at each timestep.

\begin{equation}
    \mathcal{L}_{overall} = \mathcal{L}_{task} + \delta_{p} \mathcal{L}_{ponder},
    \label{eq:final_loss}
\end{equation}

\noindent
where $\delta_{p}$ is a parameter that weights $\mathcal{L}_{ponder}$. 
\sys{} is trained to minimize $\mathcal{L}_{overall}$.

Algo.~\ref{algo:merge} presents the token merge algorithm used in \sys{}. 
A key characteristic is that tokens merged in block $\mathcal{B}^{l}$ during the first timestep are consistently merged in the same manner across subsequent timesteps within $\mathcal{B}^{l}$. 
The advantages of this approach are discussed in Sec.~\ref{subsec:eval_ablation}.

%% file: 05evaluation.tex
\section{Experiments}
~\label{sec:evaluation}





We first analyze the qualitative and quantitative results to assess how efficiently \sys{} reduces tokens for the input images (Sec.~\ref{subsec:eval_analysis}). 
Then, we conduct a comparative analysis to evaluate how effectively \sys{} reduces tokens in terms of accuracy, comparing it with existing methods, and analyze how the reduced tokens by \sys{} impact energy consumption (Sec.~\ref{subsec:eval_comparison}). 
Finally, we discuss the properties required for \sys{}'s ACT and merge to efficiently process tokens through an ablation study (Sec.~\ref{subsec:eval_ablation}).


\paragraph{Implementation details.}
We implement the simulation on Pytorch and SpikingJelly~\cite{FCD23}.
All experiments in this section are conducted on SNN-based vision transformer following Spikformer or Spikingformer architectures.
This section covers only the results of Spikformer; the results for Spikingformer are provided in the supplementary material.
We first train the Spikformer during 310 epochs and retrain during 310 epochs based on the pre-trained model.
We train the model on NVIDIA A6000 GPUs and use automatic-mixed precision (AMP)~\cite{MNA17} for training acceleration.
We set $T=4, \delta_{p}=10^{-3}, \alpha=5, \beta=-10$, and $\epsilon=0.01$.
For a fair comparison, we implemented several existing methods (e.g., TET and DT-SNN intially designed for CNN) on our target model, and these models are marked with an asterisk (*) in Tables~\ref{tab:main_exp}.
Additionally, to compare different cases, some methods were trained with a different number of timesteps than those used during inference; we marked these cases with a dagger ($\dagger$) in Table~\ref{tab:main_exp} to indicate the number of timesteps used during training. 
For example, even if the model is trained using only up to two timesteps during the training phase, it is possible to extend execution beyond two timesteps during inference. The accuracy is dependent on the timestep targeted during training.
We evaluate our method for the classification task on CIFAR-10, CIFAR-100, and TinyImageNet. 
\remove{
우리는 Pytorch와 Spikingjelly~\cite{FCD23}를 기반으로 시뮬레이션을 수행하였다.
우리는 우선 Spikformer를 기반으로 각각 310 epochs을 학습하였고, 사전학습한 모델을 활용해 우리의 방법론을 310 epochs 학습하였다.
학습 과정동안 속도 가속을 위해 automatic-mixed precision (AMP)~\cite{MNA17} acceleration를 이용했다.
$\delta_{p}=10^{-3}, \alpha=5, \beta=-10, \epsilon=0.01$으로 세팅하여 실험하였다.
우리는 우리의 방법이 일반적인 데이터셋에서 효과적임을 보이기 위해, CIFAR Dataset(CIFAR-10, CIFAR-100~\cite{KAH09}), TinyImageNet~\cite{DDS09,LYY15}에서 각각 실험하였다.
*는 pretrained model이 제공되지 않아 우리의 실험 환경에서 재측정한 정확도이다.
$^\dagger$는 학습에 사용한 timestep의 수이다.
\sys{}는 timestep을 더 줄이기 위해 기존의 timestep을 줄이는 방법론들을 적용할 수 있다.
우리는 DT-SNN과 동일하게 entrophy가 특정 threshold보다 낮아질 경우 동적으로 timestep을 중지하는 \syss{} (AT-SNN with entropy)의 결과도 함께 측정하였다.
}

\subsection{Analysis}
\label{subsec:eval_analysis}

\begin{figure}[t!]
    \centering
    \includegraphics[width=1\linewidth]{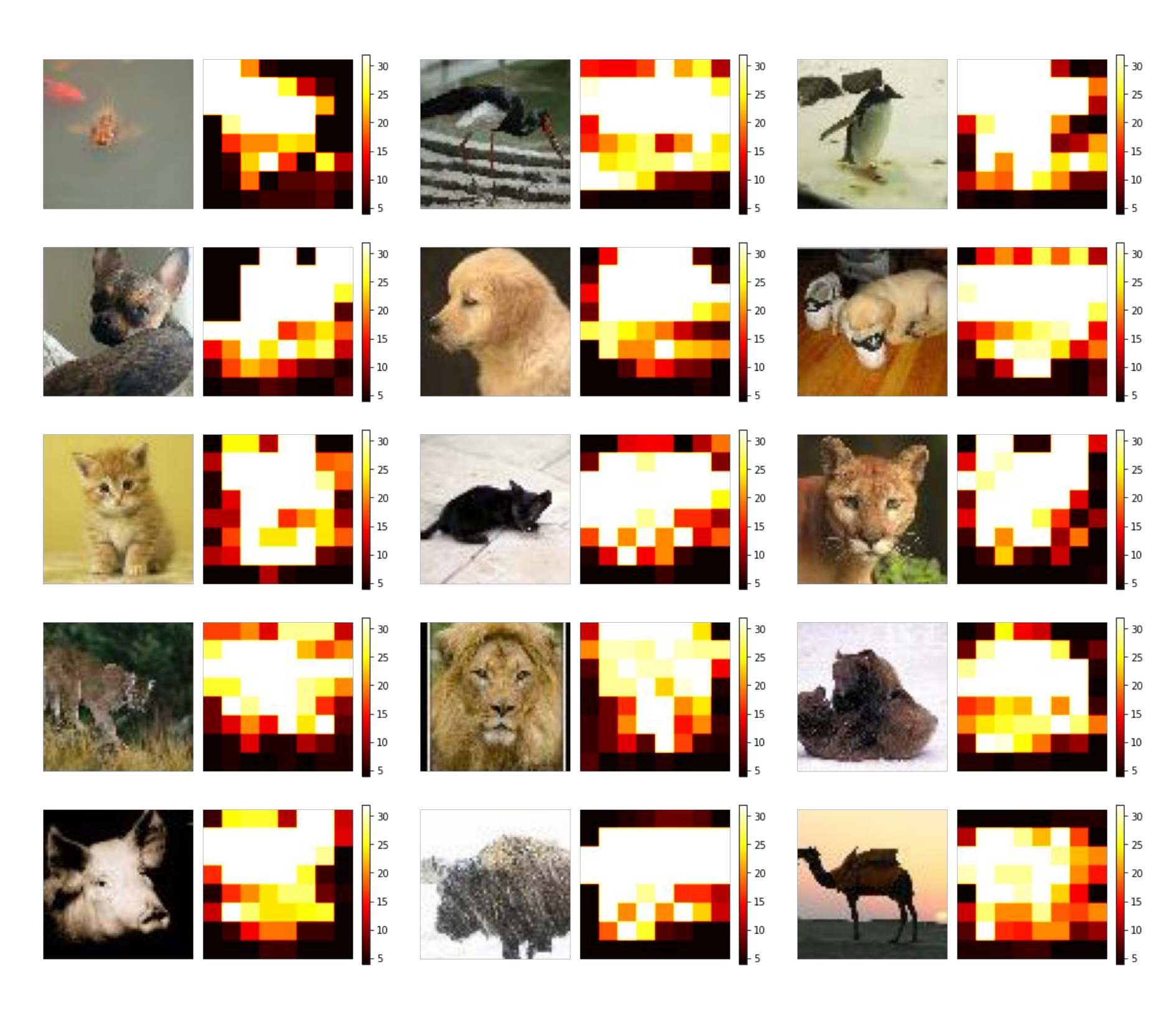}
    \caption{Original images (odd-numbered columns) and heatmaps showing the number of blocks (for four timesteps) each token processes (even-numbered columns) on TinyImageNet. Brighter colors indicate more processing per token. \sys{} halts earlier on tokens that lack visual information.}
    \label{fig:visual_samples}
     \vspace{-0.2cm}
\end{figure}


\begin{figure}[t!]
    \centering
    \includegraphics[width=1\linewidth]{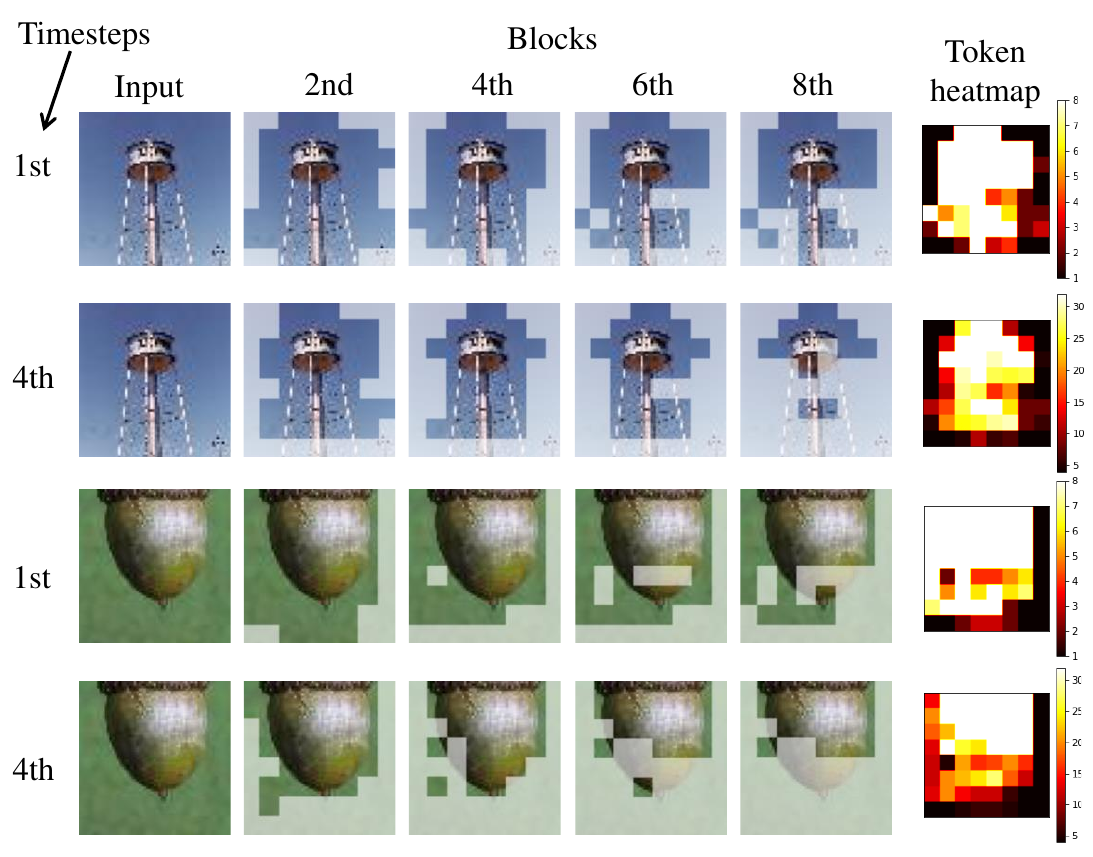}
    \caption{Example of merge and mask across different timesteps and blocks. Tokens that are masked or merged and no longer pass through inference at each block are displayed with a shaded (non-white) overlay.}
    \label{fig:timesteps_blocks_sample}
        \vspace{-0.2cm}
\end{figure}

\begin{figure}[t!]
    \centering
    \includegraphics[width=1\linewidth]{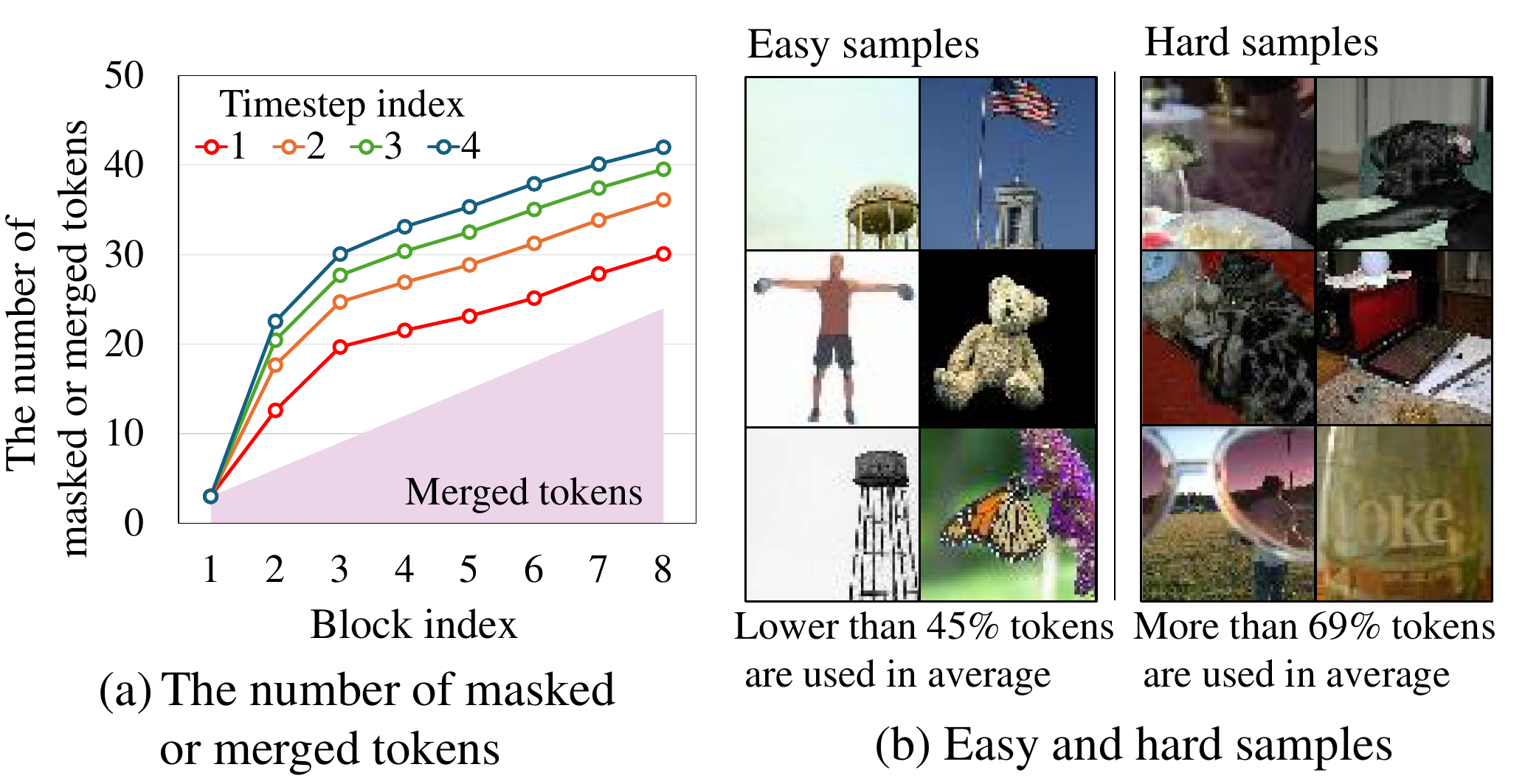}
    \caption{(a) The number of masked or merged tokens across different blocks and timesteps, and (b) visual comparison of hard and easy samples in classification.}
    \label{fig:token_sample}
        \vspace{-0.2cm}
\end{figure}

\paragraph{Qualitative results.}
For visualization of \sys{}, we use Spikformer-8-384 with eight blocks per timestep and $\gamma=3$, trained on TinyImageNet. 
Each input image contains 64 tokens ($8\times8$).
Fig.~\ref{fig:visual_samples} displays the input image (odd-numbered columns) and a heatmap (even-numbered columns) showing the number of blocks each token passes through across all timesteps. 
With four timesteps and eight blocks, the maximum processed count for each token is 32, where brighter regions indicate more processing, and darker regions indicate less (i.e., halted earlier). 
As shown in Fig.~\ref{fig:visual_samples}, \sys{} effectively reduces the number of tokens by prioritizing the removal of tokens from uninformative regions (e.g., background).
Fig.~\ref{fig:timesteps_blocks_sample} visualizes how tokens are halted over timesteps and blocks. 
Since \sys{} accumulates halting scores in two dimensions (blocks and timesteps), more tokens are halted as the block and timestep indices increase. Similar to Fig.~\ref{fig:visual_samples}, tokens from the less informative background are halted first, with an increasing number of tokens being halted over time. 

\paragraph{Quantitative results and classification difficulty.}
Fig.~\ref{fig:token_sample}(a) shows the number of tokens halted per block and timestep. 
As visualized in Fig.~\ref{fig:timesteps_blocks_sample}, more tokens are halted as the block and timestep indices increase. 
Some tokens are halted by merging (indicated by the purple region in Fig.~\ref{fig:token_sample}(a)), while the remaining tokens can be halted by ACT.
Due to the two-dimensional halting policy of \sys{}, more tokens halt as the number of timesteps increases.
Figure~\ref{fig:token_sample}(b) visualizes samples correctly classified by \sys{}, comparing those that use more tokens versus those that use fewer tokens. 
On average, easy samples utilize 45\% or fewer of all tokens per block, while hard samples use 69\% or more of all tokens per block.
We observe that \sys{} uses fewer tokens when the object in the image is clearly separated from the background and other objects. 

\begin{table}[t!]
\caption{Experiment results of SNN methods on CIFAR-10, CIFAR-100, and TinyImageNet Datasets.}
\label{tab:main_exp}
\begin{center}
\resizebox{\columnwidth}{!}{%
\begin{tabular}{|c|ccccc|}
\hline
 & Method & Architecture & Avg. tokens & T & Acc (\%) \\ \hline
\multirow{15}{*}{\rotatebox{90}{CIFAR-10}} 
    & Hybrid training             & VGG-11              & -              & 125           & 92.22       \\ \cline{2-6}
    & Diet-SNN                    & ResNet-20           & -              & 10            & 92.54       \\ \cline{2-6}
    & STBP                        & CIFARNet            & -              & 12            & 89.83       \\ \cline{2-6}
    & STBP NeuNorm                & CIFARNet            & -              & 12            & 90.53       \\ \cline{2-6}
    & TSSL-BP                     & CIFARNet            & -              & 5             & 91.41       \\ \cline{2-6}
    & STBP-tdBN                   & ResNet-19           & -              & 4             & 92.92       \\ \cline{2-6}
    & \multirow{2}{*}{TET}        & Spikformer-4-384    & $\times$1      & 2$^\dagger$   & 93.3*       \\ 
    &                             & Spikformer-4-384    & $\times$1      & 4             & 93.63*      \\ \cline{2-6}
    & \multirow{2}{*}{DT-SNN}     & Spikformer-4-384    & $\times$1      & 1.85          & 94.24*      \\
    &                             & Spikformer-4-384    & $\times$1      & 4$^\dagger$   & 94.34*      \\ \cline{2-6}
    & STATIC                      & Spikformer-4-384    & $\times$1      & 4$^\dagger$   & \textbf{\underline{94.88*}}      \\ \cline{2-6}
    & \sys{} $(\gamma=25)$        & Spikformer-4-384    & \textbf{\underline{$\times$0.28}}   & 4             & \textbf{\underline{95.06}}       \\ \cline{2-6}
    & \sys{} $(\gamma=32)$        & Spikformer-4-384    & \textbf{\underline{$\times$0.21}}   & 4             & \textbf{\underline{94.88}}       \\ \cline{2-6}
    & \syss{} $(\gamma=25)$       & Spikformer-4-384    & \textbf{\underline{$\times$0.28}}   & \textbf{\underline{2.28}}          & \textbf{\underline{94.88}}       \\ \cline{2-6}
    & \syss{} $(\gamma=32)$       & Spikformer-4-384    & \textbf{\underline{$\times$0.21}}   & \textbf{\underline{1.93}}          & \textbf{\underline{94.34}}       \\ \hline \hline

\multirow{13}{*}{\rotatebox{90}{CIFAR-100}} 
    & Hybrid training                        & VGG-11              & -              & 125           & 67.87       \\ \cline{2-6}
    & Diet-SNN                               & ResNet-20           & -              & 5             & 64.07       \\ \cline{2-6}
    & STBP-tdBN                              & ResNet-19           & -              & 4             & 70.86       \\ \cline{2-6}
    & \multirow{2}{*}{TET}                   & Spikformer-4-384    & $\times$1      & 2$^\dagger$   & 73.23*      \\ 
    &                                        & Spikformer-4-384    & $\times$1      & 4             & 74.2*       \\ \cline{2-6}
    & \multirow{2}{*}{DT-SNN}                & Spikformer-4-384    & $\times$1      & 2.35          & 75.81*      \\
    &                                        & Spikformer-4-384    & $\times$1      & 4$^\dagger$   & 76.05*      \\ \cline{2-6}
    & \multirow{2}{*}{STATIC}                & Spikformer-4-384    & $\times$1      & 2             & 75.18*      \\
    &                                        & Spikformer-4-384    & $\times$1      & 4$^\dagger$   & \textbf{\underline{77.42*}}      \\ \cline{2-6}
    & \sys{} $(\gamma=5)$                    & Spikformer-4-384    & \textbf{\underline{$\times$0.75}}   & 4             & \textbf{\underline{78.14}}       \\ \cline{2-6}
    & \sys{} $(\gamma=10)$                   & Spikformer-4-384    & \textbf{\underline{$\times$0.58}}   & 4             & \textbf{\underline{77.27}}       \\ \cline{2-6}
    & \syss{} $(\gamma=5)$                   & Spikformer-4-384    & \textbf{\underline{$\times$0.75}}   & \textbf{\underline{2.3}}           & \textbf{\underline{77.54}}       \\ \cline{2-6}
    & \syss{} $(\gamma=10)$                  & Spikformer-4-384    & \textbf{\underline{$\times$0.58}}   & \textbf{\underline{2.42}}          & \textbf{\underline{76.97}}       \\ \hline \hline

\multirow{10}{*}{\rotatebox{90}{TinyImageNet}} 
    & \multirow{2}{*}{TET}        & Spikformer-8-384    & $\times$1      & 2             & 60.6*       \\
    &                             & Spikformer-8-384    & $\times$1      & 4$^\dagger$   & 63.45*      \\ \cline{2-6}
    & \multirow{2}{*}{DT-SNN}     & Spikformer-8-384    & $\times$1      & 2.67          & 63.35*      \\ 
    &                             & Spikformer-8-384    & $\times$1      & 4$^\dagger$   & 64.01*      \\ \cline{2-6}
    & \multirow{2}{*}{STATIC}     & Spikformer-8-384    & $\times$1      & 2             & 61.27*      \\ 
    &                             & Spikformer-8-384    & $\times$1      & 4$^\dagger$   & \textbf{\underline{65.23*}}      \\ \cline{2-6}
    & \sys{} $(\gamma=3)$         & Spikformer-8-384    & \textbf{\underline{$\times$0.59}}   & 4             & \textbf{\underline{64.27}}       \\ \cline{2-6}
    & \sys{} $(\gamma=5)$         & Spikformer-8-384    & \textbf{\underline{$\times$0.45}}   & 4             & \textbf{\underline{63.67}}       \\ \cline{2-6}
    & \syss{} $(\gamma=3)$        & Spikformer-8-384    & \textbf{\underline{$\times$0.61}}   & \textbf{\underline{2.93}}          & \textbf{\underline{63.97}}       \\ \cline{2-6}
    & \syss{} $(\gamma=5)$        & Spikformer-8-384    & \textbf{\underline{$\times$0.46}}   & \textbf{\underline{3.1}}           & \textbf{\underline{63.52}}       \\ \hline
\end{tabular}%
}
\end{center}
\vspace{-0.5cm}
\end{table}

\subsection{Comparison to prior art}
\label{subsec:eval_comparison}

We consider SNN methods based on both CNNs (e.g., VGG, CIFARNet, and ResNet) and Transformers (e.g., Spikformer and Spikingformer). Among the methods considered, except for TET, DT-SNN, STATIC (i.e., a Transformer model without any lightweight techniques), \sys{}, and \syss{}, all are based on CNN models. For the methods based on CNN models, we used the performance data reported in the original papers without modification~\cite{RSP20,WDL18,ZWL20}.
We further explored \syss{}, an extension of \sys{}. \syss{} (\sys{} with entropy) follows the DT-SNN approach, halting computation when the entropy of confidence scores exceeds a threshold.


\paragraph{Accuracy and token usage ratio.} 
As shown in Table~\ref{tab:main_exp}, Transformer-based methods outperform CNN-based ones with fewer timesteps and higher accuracy. 
For CIFAR-10, we set $\gamma=25$ and $\gamma=32$, and for CIFAR-100, $\gamma=5$ and $\gamma=10$, providing a trade-off between accuracy and the number of processed tokens.
Compared to STATIC, which achieves the highest accuracy among existing methods (94.88\% for CIFAR-10 and 77.42\% for CIFAR-100), \sys{} with $\gamma=25$ (for CIFAR-10) and $\gamma=5$ (for CIFAR-100) achieves even higher accuracy while using only 28\% and 75\% of the tokens, respectively. 
For TinyImageNet, we experimented with Spikformer-8-384, using 64 tokens per block, eight blocks, and an embedding dimension of 384, comparing accuracy and token usage at $\gamma=3$ and $\gamma=5$. At $\gamma=3$, \sys{} achieves 64.27\% accuracy while using 0.59 times fewer tokens per block. 
If \sys{} uses $0.59\times4=2.36$ tokens across all timesteps, it surpasses TET and STATIC at two timesteps ($1\times2$) and DT-SNN at 2.67 timesteps ($1\times2.67$) in accuracy. 
Although \sys{} has slightly lower accuracy than STATIC (65.23\%), it dramatically reduces token usage, making it highly competitive in terms of energy consumption (discussed in the next paragraph).
Additionally, \syss{} significantly reduces token usage by lowering the number of timesteps, with only a minor decrease in accuracy.

\paragraph{Energy consumption.}
We measured the energy consumption of each module by calculating the number of operations in Spikformer during the inference phase. 
Following the widely accepted measurement methods in previous SNN studies~\cite{ZZH22,ZYZ23,HOM14}, the equation for calculating energy consumption is provided in the supplement. Fig.~\ref{fig:energy_consumption} illustrates the power consumption measured on CIFAR-10, CIFAR-100, and TinyImageNet, respectively. 
To demonstrate that \sys{} exhibits lower energy consumption despite achieving higher accuracy, we compared it to STATIC on TinyImageNet by reducing the accuracy with the timestep set to three.
As shown in Fig.~\ref{fig:energy_consumption}(c), \sys{} cannot reduce power consumption in SPS since it adaptively executes tokens only in the encoder block $\mathcal{B}$. 
Nevertheless, \sys{} significantly reduces energy consumption compared to methods other than STATIC on TinyImageNet, and it also exhibits lower energy consumption than STATIC on TinyImageNet.

\begin{figure}[t!]
    \centering
    \includegraphics[width=\linewidth]{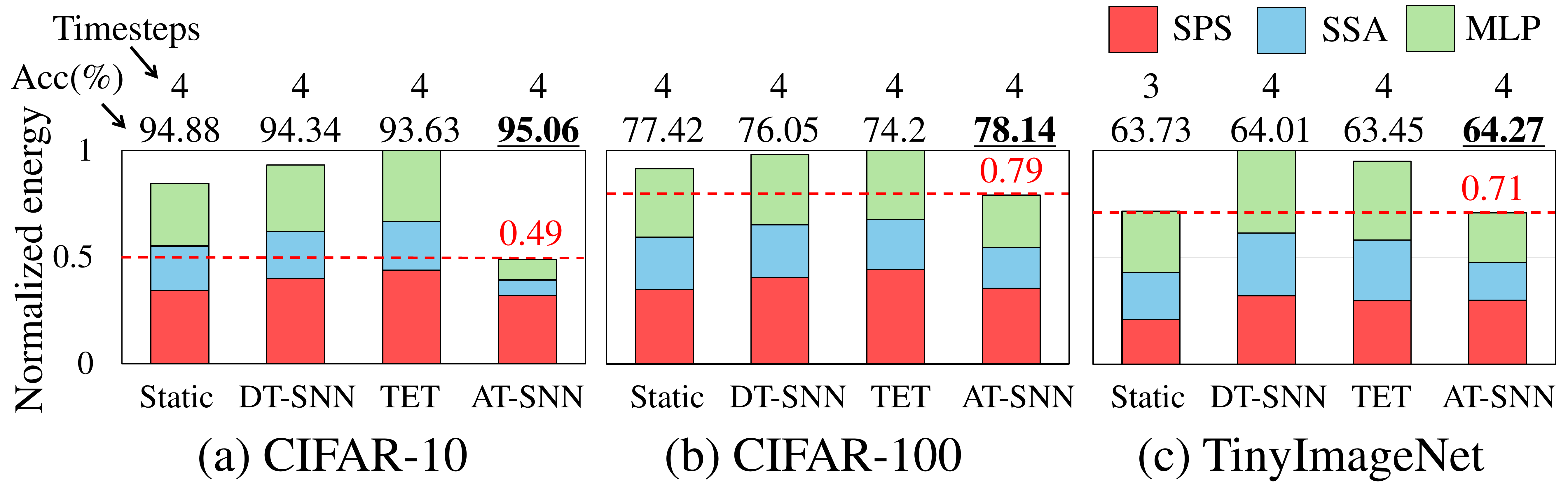}
    \caption{Energy consumption comparison on (a) CIFAR-10, (b) CIFAR-100, and (c) TinyImageNet.}
    \label{fig:energy_consumption}
       \vspace{-0.2cm}
\end{figure}



\subsection{Ablation Study}
\label{subsec:eval_ablation}

\begin{figure}[t!]
    \centering
    \includegraphics[width=\linewidth]{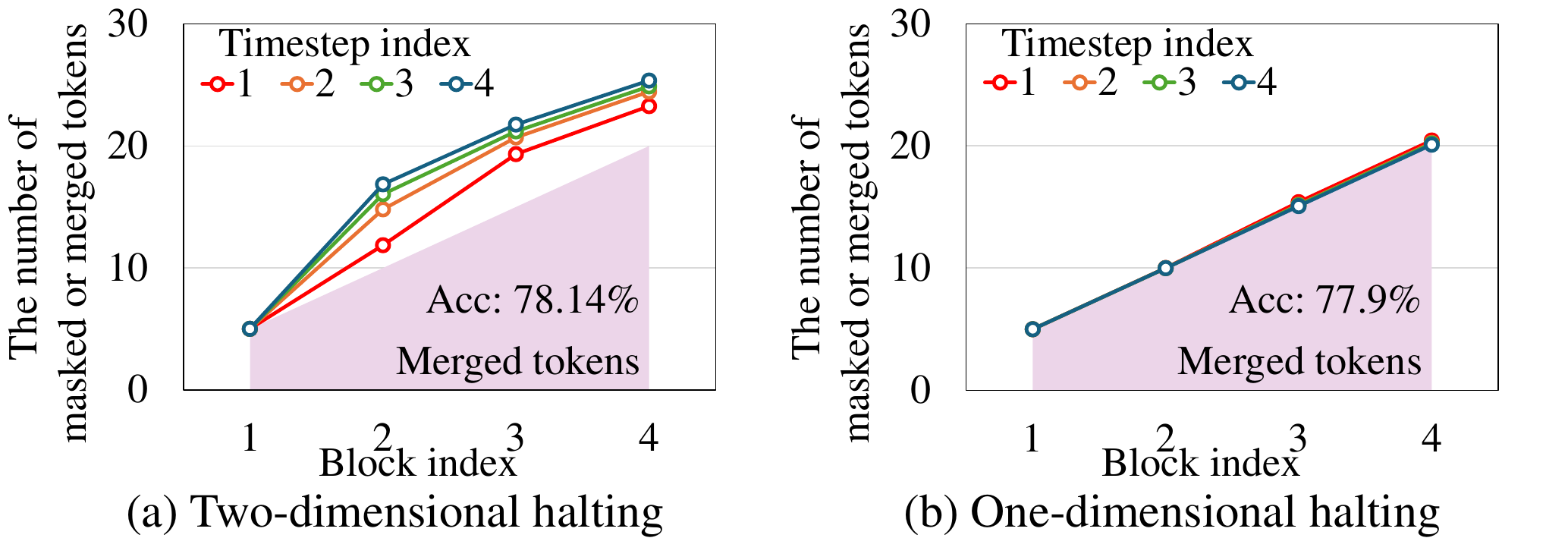}
    \caption{Comparison between two-dimensional and one-dimensional halting on CIFAR-100. The former follows Eq.~\eqref{eq:M_score}, while the latter prevents halting score accumulation across timesteps.}
    \label{fig:dimensional_AT_SNN}
        \vspace{-0.2cm}
\end{figure}

\paragraph{Two- vs one-dimensional halting.}
Fig.~\ref{fig:dimensional_AT_SNN} compares the halting score accumulation methods on CIFAR-100: one that accumulates scores across two dimensions (both timestep and block-levels as per Eq.~\eqref{eq:M_score}) and another that accumulates only across one dimension (block-level only). 
As shown in Fig.~\ref{fig:dimensional_AT_SNN}, the two-dimensional halting mechanism achieves higher accuracy while removing more tokens compared to the one-dimensional halting. 
This is because, by definition, the LHS of Eq.~\eqref{eq:nb} becomes larger under two-dimensional halting than under one-dimensional halting, which in turn increases the LHS of Eq.~\eqref{eq:ponder_loss}, leading to more tokens being halted. 
Moreover, two-dimensional halting achieves even higher accuracy than one-dimensional halting, a result that is related to the characteristics of spiking neurons discussed in the following paragraph.

\paragraph{Temporal-awareness of merge and halting.}

\begin{figure}[t!]
    \centering
    \includegraphics[width=1\linewidth]{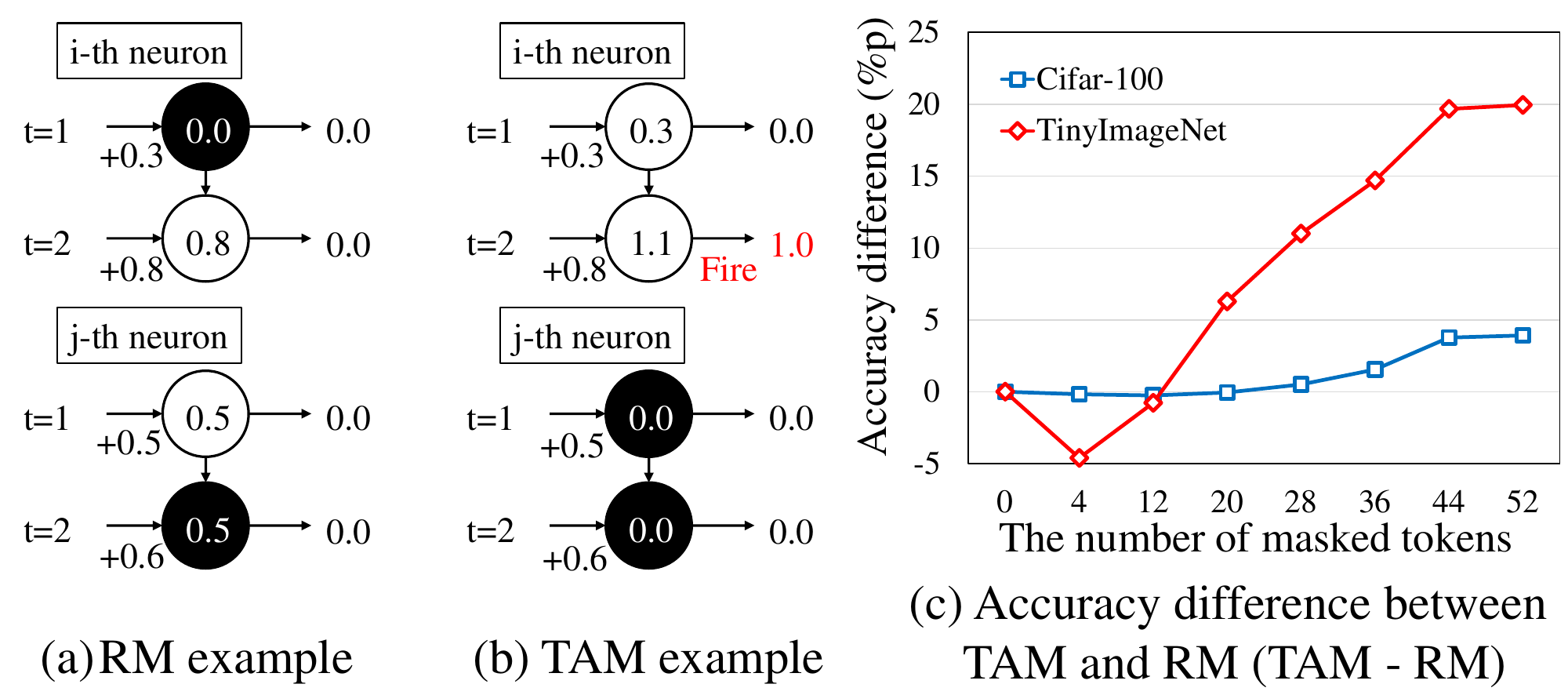}
    \caption{Comparison between random masking and temporal aware masking.}
    \label{fig:TAM}
        \vspace{-0.2cm}
\end{figure}

In SNNs, spiking neurons accumulate inputs over multiple timesteps and fire when their value exceeds a certain threshold. 
We observed that temporally-aware masking (TAM), where the same neurons are masked across multiple timesteps, reduces accuracy less than random masking (RM). For instance, as illustrated in Fig.~\ref{fig:TAM}(a) and (b), consider $i$-th and $j$-th neurons receiving inputs of (0.3, 0.8) and (0.5, 0.6) over two timesteps. 
If we randomly mask two out of the four inputs, no neuron may fire, as in Fig.~\ref{fig:TAM}(a). 
However, by consistently masking only the $j$-th neuron, the $i$-th neuron can fire and propagate information, as shown in Fig.~\ref{fig:TAM}(b). 
To explore this further, we increased the number of masked tokens and compared the performance of TAM and RM. 
Fig.~\ref{fig:TAM}(c) shows the difference in accuracy between TAM and RM (accuracy of TAM $-$ accuracy of RM) as the number of masked tokens increases. 
While RM achieves higher accuracy with fewer masked tokens, TAM outperforms RM as the number of masked tokens increases. 
Based on this observation, we designed Algorithm~\ref{algo:merge} in \sys{} to ensure that the same tokens are consistently merged across timesteps. 
This also explains why two-dimensional halting, which consistently halts the same tokens, achieves higher accuracy than one-dimensional halting, as shown in Fig.~\ref{fig:dimensional_AT_SNN}.

%% file: 06conclusion.tex
\section{Conclusion}
\label{sec:conclusion}

In this paper, we introduced \sys{}, a framework designed to dynamically adjust the number of tokens processed during inference in directly trained SNN-based ViTs, with the aim of optimizing power consumption. 
We extended ACT mechanism, traditionally applied to RNNs and ViTs, to selectively discard less informative spatial tokens in SNN-based ViTs. 
Furthermore, we proposed a token-merge mechanism based on token similarity, which effectively reduced the token count while enhancing accuracy. 
We implemented \sys{} on Spikformer and demonstrated its effectiveness in achieving superior energy efficiency and accuracy on image classification tasks, including CIFAR-10, CIFAR-100, and TinyImageNet, compared to state-of-the-art methods.

\paragraph{Limitation.}
As discussed in Fig.~\ref{fig:energy_consumption}, \sys{} aims to reduce the number of tokens involved in inference, and thus does not reduce energy consumption in the SPS. 
Reducing energy consumption in the SPS is a subject for future work.